\documentclass[10pt,conference]{IEEEtran}
\usepackage{cite}
\usepackage[tight,footnotesize]{subfigure}
\usepackage{color}
\usepackage[normalem]{ulem}

\ifCLASSINFOpdf
   \usepackage[pdftex]{graphicx}
\else
\fi
\usepackage[cmex10]{amsmath}
\usepackage{amssymb}
\usepackage[ruled]{algorithm2e}
\usepackage[singlelinecheck=off]{caption}

\usepackage{url}

\begin{document}

\title{SafeRNet: Safe Transportation Routing in the era of Internet of Vehicles and Mobile Crowd Sensing}

 
\author{\IEEEauthorblockN{Qun Liu\IEEEauthorrefmark{1},
 Suman Kumar\IEEEauthorrefmark{1}, Vijay Mago\IEEEauthorrefmark{2}}
   \IEEEauthorblockA{\IEEEauthorrefmark{1}Department of Computer Science, Troy University, Troy, AL, USA}
 \IEEEauthorblockA{\IEEEauthorrefmark{2}Department of Computer Science, Lakehead University, Thunder Bay, ON, Canada\\
Email: \{qliu120075, skumar\}@troy.edu, vmago@lakeheadu.ca
 }}

\maketitle

%


\begin{abstract}
World wide road traffic fatality and accident rates are high, and this is true even in technologically advanced countries like the USA. Despite the advances in Intelligent Trans- portation Systems, safe transportation routing i.e., finding safest routes is largely an overlooked paradigm. In recent years, large amount of traffic data has been produced by people, Internet of Vehicles and Internet of Things (IoT). Also, thanks to advances in cloud computing and proliferation of mobile communication technologies, it is now possible to perform analysis on vast amount of generated data (crowd sourced) and deliver the result back to users in real time. This paper proposes SafeRNet, a safe route computation framework which takes advantage of these technologies to analyze streaming traffic data and historical data to effectively infer safe routes and deliver them back to users in real time. SafeRNet utilizes Bayesian network to formulate safe route model. Furthermore, a case study is presented to demonstrate the effectiveness of our approach using real traffic data. SafeRNet intends to improve drivers’ safety in a modern technology rich transportation system.


\end{abstract}


%
\IEEEpeerreviewmaketitle

\section{Introduction and Motivation}
\label{intro}
World Health Organization's 2015 Global Status report indicates that the total number of road traffic deaths are 1.25 million per year world wide. Poor countries that lack good infrastructure reported the traffic fatality rate to be at 24.1 per 100,000, where as in technology rich high income countries the same is observed to be 9.2 per 100,000. According to current U.S. Department of Transportation statistics, \emph{``There were 29,989 fatal motor vehicle crashes in the United States in 2014 in which 32,675 deaths occurred. This resulted in 10.2 deaths per 100,000 people and 1.08 deaths per 100 million vehicle miles traveled.\footnote{ Federal Highway Administration. 2015. Highway statistics, 2014. Washington, DC: U.S. Department of Transportation, http://www.iihs.org/iihs/topics/t/general-statistics/fatalityfacts/state-by-state-overview }''} In USA, same study shows fatality rate ranges from 3.5 to 25.7 per 100,000 people where as death rate ranges from 0.57 to 1.65 per 100 million vehicle miles traveled~\cite{WhoRoadSafety2015}. Although it is extremely difficult to bring the fatality rates down to zero, nevertheless, these statistics are startling, especially, in technologically advanced countries. We believe real time safe routing i.e., computing and delivery of safest route to end users can address the safety related problems to great extent and thus help reduce the fatality as well as accident rates. We argue that safe routing problem can be addressed effectively and efficiently by using emerging technologies like Internet of Vehicle (IoV) and Mobile Crowd Sourcing as we explain in later paragraphs.

Intelligent transportation has come a long way in past couple of decades, and Internet of Vehicle (IoV) is adding a new dimension to it. Because of computing and communications capabilities, IoV has a potential to become the corner stone in delivering and consuming rich applications in safe and secure manner. IoV enables gathering and sharing information about the traffic, road, and vehicle itself by using V2V (vehicle-to-vehicle), V2H (Vehicle-to-Human), V2S (Vehicle-to-Sensor) communications and interactions. This brings us to the question: what role IoV can play in guiding and supervising vehicles to help improve safety in transportation system? Our proposal is an attempt to answer this question.

Recently, the Mobile crowd sensing (MCS), a new sensing paradigm~\cite{guo2014participatory} is producing a lot of useful traffic data, such as vehicle trajectory and lane changing behavior~\cite{ruj2011data}. The data produced and collected through mobile phones is delivered to the cloud for processing purposes. Unfortunately, a large amount of valuable traffic data has not been effectively utilized in addressing the safety issues on the road. What role these data would play in safe route planning is not clear. Our proposed work puts a great emphasis on utilizing the dynamic traffic data in a real time fashion. 
\begin{figure*}[!t]
\centering
\includegraphics[scale=1]{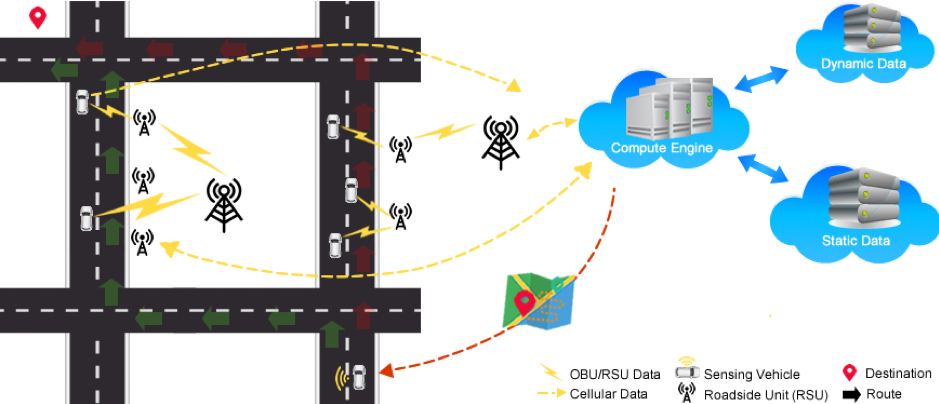}
\caption{High Level System Architecture Showing Flow of Information}
\label{general}
\end{figure*}
We propose a SafeRNet, Figure \ref{general}, a framework which addresses the gap in research focusing on big data generated by MCS paradigm, and by vehicles and IoT deployed on road side to effectively infer safe routes. SafeRNet aims to show what a safe route solution would look like in an IoV rich world where, because of advances in cloud computing, analyzing the MCS data has become a reality.

To the best of our knowledge this is the first work of its kind. We summarize our contributions as below:
\begin{itemize}
\item A novel data flow framework integrating MCS, IoV and cloud computing is proposed showing user-to-cloud and cloud-to-user interaction. The framework combines real time MCS data, IoV data and historical data to deliver safest route to the end users.

\item Notion of safety is conceptualized by using Bayesian network model.

\item Safest route computation is formulated and a possible solution is presented.

\item A case study is presented to demonstrate the feasibility and effectiveness of proposed framework.
\end{itemize}



The paper is organized as follows: Section \ref{Related} presents background and related work briefly. SafeRNet architecture, Bayesian network modeling approach for safety probability estimation and safe routing problem formulations are presented in Section~\ref{System}. Section \ref{experiment} demonstrates experimental result. Finally, \ref{conclusion} presents conclusion and future work.


\section{Background and Related Work}
\label{Related}
Mobile Crowd Sensing is a new sensing paradigm that has an advantage of large scale sensing as compared to traditional approaches. No doubt it has become popular and therefore, a reality \cite{guo2014participatory}  as it enables high sensing accuracy with very low error rate\cite{guo2013mobile}. MCS coupled with a large variety of devices has a potential to become a cutting-edge technology for Internet of Things that would provide a seamless sensor data transfer via Internet. There are already many applications appearing in literature \cite{ganti2011mobile} that use MCS data gathering paradigm. 

Internet of Vehicles is a rapidly developing communication paradigm \cite{kolls2006communication} that possesses the ability to perform accurate positioning even in the blocked GPS signal scenario \cite{prinsloo2016accurate}. In the typical setting of vehicular ad hoc networks, it has gained attention from researches world wide to address vehicle collision warning problem and traffic information dissemination issues \cite{eze2016advances}. This sensing paradigm has a potential to bring enormous attention on the related area, such as traffic prediction, which has a potential to make real-time performance better \cite{wan2016mobile}. Machine learning has been heavily used in ITS \cite{lepine2016use} in mining mobile data stream~\cite{krishnaswamy2012mobile}, in constructing mathematical models\cite{Kotkowski2015} and for exchange of information among vehicles\cite{fogue2016non}. Bayesian network is adopted in our research because of its ability to concisely represent probabilistic relationships \cite{cooper1990computational} and its previous successful application \cite{belyi2015multi}. However, other methods can also be utilized to analyze data and perform predictions in our framework. 


\section{SafeRNet: Architecture, Design and Modeling}
\label{System}
In this section, notion of safety used in our work is explained. Then, we present an overview of our proposed system. Furthermore, we describe Bayesian network modeling approach to compute safety probability (described later) and explain how safe route formulation utilizes this information to compute safest route.

\subsection{Notion of Safety}
In our proposed work, road traffic safety is defined as a way to measure traffic fatality rate, accident rates and also near collision incidents. We believe that safety issue can arise mainly because of three reasons, {\it law compliance, road condition and hazardous behavior}. Non-adherence to law related safety issue arises when drivers break the law such as running the traffic lights, going over/under legal speed limit, etc. Road condition is related to light intensity on the road, type of road (paved/non-paved), road lanes, weather condition, time of day, day of week etc. Hazardous behavior related to reckless driving such as frequent lane changing, light flashing, honking etc. There are other factors that contribute to the accidents such as presence of pedestrians on the road, driver's mental state etc. There can be many factors that impact traffic safety, however, we limit our study to the data that can be gathered by using IoV with MCS data gathering paradigm. Nevertheless, the distinction among causes of accidents are important to take preventive and cautionary action. For example, the extent of law compliance may trigger a proactive government intervention to bring order to a road. 

 The safety probability, $p(s_i)$, of a road $i$ is defined as below:
\begin{equation}
p(s_i) = 1- p(c_i)
\label{eq:}
\end{equation}
 where $p(c_i)$ is collision probability 
It is to be noted that for simplicity we focus on collision only and not on number of causalities or number of vehicles involved in a collision.    

\subsection{System Overview}
\label{sysarch}
End-to-end data flow in SafeRNet architecture is shown in Fig. \ref{general}. SafeRnet can be broadly viewed as the integration of three modules: (i) sensing and communication; (ii) databases for storing dynamic as well as historical data; and (iii) compute engine to analyze the data. The functional details of these modules are presented next.

\subsubsection{Sensing and Communication}
IoV coupled with deployed sensing and communication technologies on the road act as a data source and communication mediums.  In our IoV model, vehicles act as mobile sensor nodes that are equipped with the On-Board Units (OBU) which can communicate to Road Side Units (RSUs)/OBU or directly to the cloud via cellular network. The RSUs are fixed nodes that serve as an infrastructure to facilitate data communication to/from remote cloud. Deployed sensors along the road such as cameras, speed detectors, etc. acts as data acquisition system that obtain and send additional traffic data on surrounding area to the cloud. 

\subsubsection{Databases}
Collected data is associated with a particular road or road segments. Data is classified in three broad categories - law compliance, road condition and hazardous behavior. These three categories are related to safety issues as described in the previous section. The proposed classification has a potential to be used to provide a preference based safe routes to end-users. The proposed architecture uses two kinds of databases, one for storing the dynamic data and other for static data. The data that does not change in small time frame are static data, for example, road type, road zone, map data are static data. Dynamic data are either streaming traffic data or data that changes in short time frame. For example, data such as weather condition, light condition sometimes changes frequently and therefore they belong to dynamic data category. Also, streaming data such as vehicle density, lane changing behavior, speed of vehicle are also dynamic data and are stored in dynamic database. We use \textbf{\it short time frame} as a more generic term with time units: minute(s), hour(s) or a day.

\subsubsection{Compute Engine}
\begin{figure}[h]
  \includegraphics[scale=0.45]{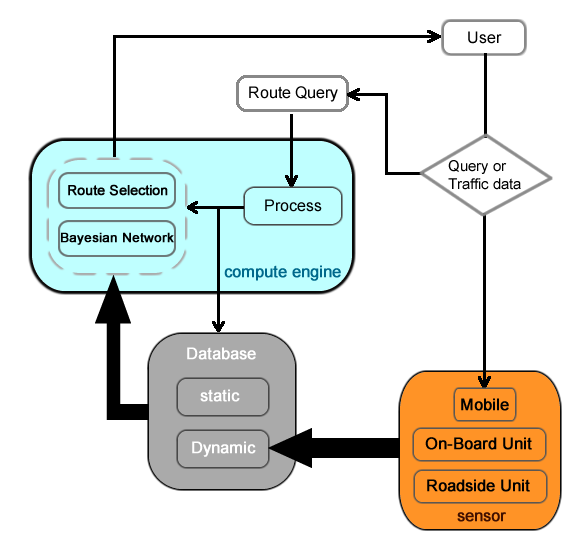}
	\centering
  \caption{Functional block diagram showing processing of information in various modules SafeRNet framework}
  \label{flowchart}
\end{figure}

Compute engine uses Bayesian network model to compute safety probability for road/road segment by referring to historical data (see Figure~\ref{flowchart}) and route selection component to compute safest route. It is to be noted that compute engine has access to both types of data and it has the ability to update its database when an event of interest is detected. In our framework, Bayesian network periodically updates its structure adapting to new data sources and attributes. Because of its ability to learn causal relationships, Bayesian network becomes a clear choice to serve as a compute engine in our work. Also, desired properties like handling of incomplete data, prevention of over fitting of data and straight forward construction of prior knowledge make Bayesian network an excellent choice for our work. Once the Bayesian network is constructed, the road segments are assigned safety probability in the road network which is a part of map data. Furthermore, the safest route is computed on the graph obtained by map data with edges associated with safety probability (see Section~\ref{selection}).

\subsection{Bayesian Network Modeling}
\label{model}
Bayesian network (BN) also known as Bayesian Belief network, is a probabilistic directed acyclic graphical model which represents the conditional independence relationships. The conditional independence is capable of probabilistic representation and reasoning among variables by using a directed acyclic graph (DAG) \cite{belyi2015multi}. 
The nodes and directed edges of BN are random variables and conditional dependence relationships among variables, respectively. Nodes are conditionally dependent given the value of their parents. For each node in the Bayesian network there is a conditional probability table (CPT) that serves as prior probabilities for BN. These prior probabilities are used to compute total joint probability according to below equation:
\begin{equation}
 P(X_{1},...,X_{n})=\prod_{i=1}^nP(X_{i}\mid Pa(X_{i}))
\label{eq:BNtj}
\end{equation}

where $X_{i}$ is the ${i}$th node in the set of n nodes of the network and $Pa(X_{i})$ denotes the parent node of node $X_{i}$. The aim of BN learning is to support the training data by finding the detailed relationship among variables as well as their corresponding CPTs. Next, algorithms for BN scoring metrics, structure learning and parameter learning are discussed.

\subsubsection{Scoring}
Scoring metrics is a method to measure the performance and quality of the network for a given set of data. The Bayesian metric \cite{ouali2006data} for a specific Bayesian network structure $B$ for a database $D$ is defined as:
\begin{equation}
 Q(B,D)=P(B)\prod_{i=0}^n\prod_{j=1}^{q_{i}}\frac{\Gamma(N_{ij}^{'})}{\Gamma(N_{ij}^{'}+N_{ij})}\prod_{k=1}^{r_{i}}\frac{\Gamma(N_{ijk}^{'}+N_{ijk}))}{\Gamma(N_{ijk}^{'})}
\label{eq:Bm}
\end{equation}
\noindent where $P(B)$ is the prior on $B$, $r_{i}$ ($1\leq i \leq n$) is the cardinality of $X_{i}$ variables, $N_{ij}$ represents the number of records in the database which $pa(X_{i})$ takes its ${j}$th value, $N_{ijk}$ represents the number of records in the database which $pa(X_{i})$ takes its ${j}$th value and which $X_{i}$ takes its ${k}$th value. $N_{ij}^{'}$ and $N_{ijk}^{'}$ are the prior on $N_{ij}$ and $N_{ijk}$, the gamma-function $N_{ij}^{'}$ and $N_{ijk}^{'}$ represent the choices of priors on counts restricted by $N_{ij}^{'}=\sum_{k=1}^{r_{i}}N_{ijk}^{'}$. When $N_{ijk}^{'}$ assigned 1, the K2 metric is obtained.

\subsubsection{Structure Learning}
To learn the best structure of Bayesian network we adopt K2 algorithm \cite{cooper1992bayesian}, a greedy algorithm, in our this study. For a given database $D$, prior knowledge $\sigma$, structure learning aim to find an optimal structure $B_s$ with the best score as per below equation:
\begin{equation}
 B_s\leftarrow argmaxP(B\mid D,\sigma)
\label{eq:BBN}
\end{equation}

\subsubsection{Parameter Learning}
Once the Bayesian structure is constructed, one can build the conditional probability table (CPT) for each relationship of the nodes. Given the database $D$ and node $\theta$, we have:
\begin{equation}
 P(\theta\mid D)=\frac{P(\theta)P(D\mid\theta)}{P(D)}=\frac{\prod_{k=1}^r\theta_k^{\alpha k+N_{k}-1}\Gamma(\alpha+N)}{\prod_{k=1}^r\Gamma(\alpha_{k}+N_{k})}
\label{eq:BNP}
\end{equation}
where $\alpha=\sum_{k=1}^r\alpha_{k}$ and $\alpha_{k}>0$, $\alpha_{(.)}$ are Hyperparameters representing the Dirichlet priors \cite{heckerman1995learning} which is the probability distribution for prior knowledge of the relationships among variables in $D$.

\subsubsection{BN Inference}
Marginal probability is calculated by using the above structure and learned CPTs for each node from the observations in the Bayesian network for each road segments. Furthermore, we assign those probabilities to edges in the road network graph obtained from map data.

\subsection{Safe route problem formulation}
\label{selection}
In this section, we present our model of safe route and further safest route formulation is presented and its transformation to shortest path problem is explained. 
%

\subsubsection{Safe Route Model}
In our work, safety probability of road segments are computed by using Bayesian network as described in previous subsection. For simplicity sake, we assume that safety probability of roads are independent of each other, therefore, following law of independence, safety probability of a route is defined as $p(R)$ below:

\begin{equation}
p(R) = {\displaystyle \prod_{i \in R}^{} p_{i}}
\label{safetyroute}
\end{equation}

where $R$ is the set of roads forming a route, $i$ is some road in route R and $p_i$ is corresponding safety probability.

%
%
%
%
 
\begin{figure}[ht]
\centering
\includegraphics[scale=0.4]{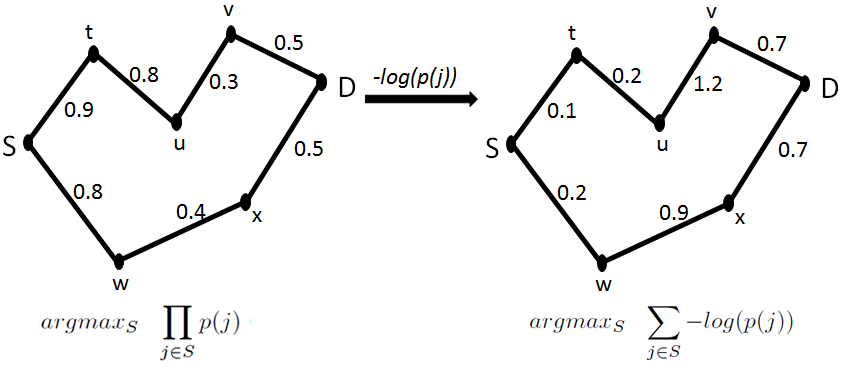}
\caption{Example Graph showing transformation of maximization problem of multiplied weights of a graph into a shortest path problem.}
\label{problemtransform}
\end{figure}
\subsubsection{Safest Route}

From a source to a destination, there exists a set of roads $S$ forming set of routes $R$. Following the safe route model described in previous section, the safest route $R_s$ is given by following formulation:
\begin{equation}
argmax_{R}\ \ p(i)
\label{safestroute1}
\end{equation}
where $i$ $\in$ $R$.

The safest route $R_s$ is the result of a set of roads that gives maximum safety route probability, therefore, following ~\ref{safetyroute} 
 the problem of finding safest route can be formulated as below:
\begin{equation}
argmax_{S}\ \ \prod_{j \in S}^{} p(j)
\label{safestroute1}
\end{equation}

Since we are interested in finding the safest route $R_s$ the problem can be reformulated as below:

\begin{equation}
argmax_{S}\ \ log(\prod_{j \in S}^{} p(j))
\label{safestroute1}
\end{equation}
 
Law of logarithms transforms the maximization of multiplication problem into maximization of summation problem:

\begin{equation}
argmax_{S}\ \ \sum_{j \in S}^{} log(p(j))
\label{safestroute1}
\end{equation}

Furthermore, since $0 \leq p(j) \leq$ log(p(j)) is a negative quantity, after applying unary operator $-$, the problem becomes a minimization problem:

\begin{equation}
argmin_{S}\ \ \sum_{j \in S}^{} -log(p(j))
\label{safestroute1}
\end{equation}

Above equation indicates, to find the safest route, we must find the shortest path in a graph where $-log(p(i))$ is the weight of road $j$ ( $\in S$). Such problem can be solved by using Dijskstra's shortest path algorithm in $O(|E|+|V|log|V|)$ using a Fibonacci heap. Example of transformation is shown in Fig.~\ref{problemtransform}.

Safest route score is defined as below:

\begin{equation}
-ln(1 - p(R_s))
\label{safestroute1}
\end{equation}

therefore, higher safety score means safer route. 

%

\begin{table*}[t]
\caption{Description of the variables used in case study}
\label{DV}
\begin{tabular}{lll}
\hline
Variable & Description            & Discrete state                                                                                                                                                       \\ \hline
TR       & Type of road           & 0 highway, 1 district or province road                                                                                                                             \\
TRL      & Type of road lanes     & 0 road with one road lane, 1 road with separated road lanes                                                                 \\
RF       & Road factors           & 0 bad road surface, 1 faulty signals, 2 faulty lighting, 3 road works, 4 queue, 5 downhill, 6 curve,7 bad visibility\\
WC       & Weather conditions     & 0 normal weather, 1 rain, 2 fog, 3 wind, 4 snow, 5 hail, 6 other weather                                                     \\
RC       & Road conditions        &  0 dry road surface, 1 wet road surface, 2 snow on road surface, 3 clean road surface, 4 dirty road surface                  \\
LC       & Light conditions       &  0 daylight, 1 twilight, 2 public lighting, 3 night                                                                           \\
W        & Week                   &  0 week, 1 weekend                                                                                                                                                     \\
PD       & Part of the day        &  0 morn. rush hour\_9h, 1 morn.10-12h, 2 noon13\_15h, 3 eve. rush hour16\_18h, 4 eve.19\_21h, 5 night22\_6h  \\
C        & Collision              &  0 none, 1 collision                                                                                                                                                   \\
V        & Velocity               &  0 Low, 1 Normal, 2 High                                                                                                                                               \\
VD       & Vehicle Density        &  0 low, 1 high                                                                                                                                                         \\
LCB      & Lane Changing Behavior &  0 not frequent, 1 frequent                                                                                                                                            \\
RZ       & Road Zone              &  0 none, 1 commercial, 2 residential                                                                                                                                                                                                                                                                                 \\ \hline
\end{tabular}
\end{table*}

\section{Experiment}
\label{experiment}
In this section, we demonstrate the working and effectiveness of SafeRNet. We used real traffic data to build Bayesian network and present a case study.
\subsection{Bayesian Network Structure Learning}
%

\begin{figure}[h]
  \includegraphics[scale=1]{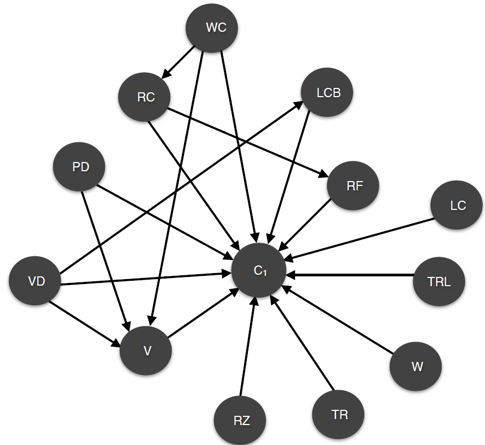}
	\centering
  \caption{Bayesian network structure for road safety probability estimation.}
  \label{BNstructure}
\end{figure}

We use the dataset obtained from {\it Frequent Itemset Mining Dataset Repository}\footnote{http://fimi.ua.ac.be/data/} research community. Further details about the dataset can be found here~\cite{geurts2003}. The missing data attribute values are generated by using Gaussian distribution function. A total of 160k records are used in our study. The data attributes used in our study are listed in Table~\ref{DV}. Furthermore, Bayesian network structure is trained by using 80\% of the sample data set and rest of the data is used for testing. The obtained Bayesian network structure is shown in Figure~\ref{BNstructure}.

\begin{figure}[h] 
  \centering
  \subfigure[The route map (Dothan to Atlanta)]{\label{routemap}
   \includegraphics[width=0.23\textwidth]{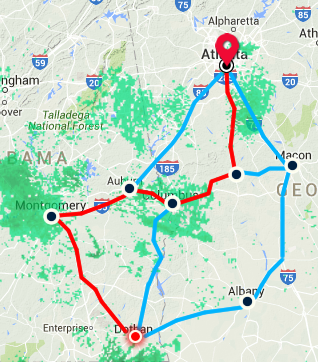}} 
  \subfigure[Extracted road network graph]{\label{shortframe}
   \includegraphics[width=0.23\textwidth]{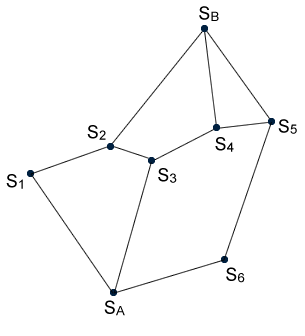}} 
   \caption{Map of geographical segment and the extracted road network used in case study }
  \label{shortframe}
\end{figure}

\subsection{Case Study}
\begin{figure*}[ht]
  \includegraphics{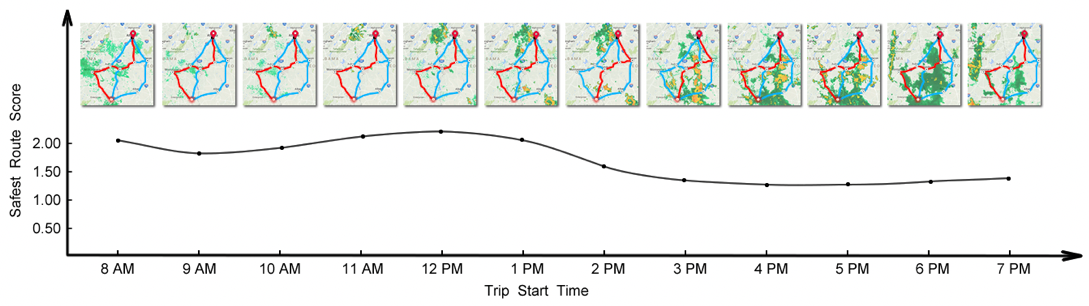}
	\centering
  \caption{Safest routes with its associated safest routes score for different trip start time}
  \label{routesmap}
\end{figure*}

To study the effectiveness of our proposed model and make the simulated scenarios as close to reality as possible. We consider a long distance route scenario that spans a geography where there is a visible and considerable variation of certain spatial data. For demonstration purposes, we focus on a geographical region between Dothan, Alabama and Atlanta, Georgia. The roads between cities are road segments. Figure~\ref{routemap} shows the map data and extracted road network graph is shown in Figure~\ref{shortframe}. The presented scenario utilizes the real weather data from radar map provided by {\it WunderMap}\footnote{https://www.wunderground.com/wundermap/} on June 26th, 2016 for geography shown in figure~\ref{shortframe}. The weather map shows a moving storm in that geography on that day. The weather data is recorded periodically at a set interval of 1 hour starting from 8AM to 7PM. Weather impacts the safety probability of the road segments shown in Figure \ref{shortframe}.

Figure \ref{routesmap} shows the variation of safety route score for different trip start time. Trip is defined as a set of routes between Dothan, Alabama and Atlanta, Georgia. Figure shows that safest route varies based on weather conditions in that region. For example, one would take a different route if the trip is started at 1PM as compared to the route if the trip is started at 2PM. Also, the safest route may be not desirable if the safest route score is low (lower score means lesser safety, see equation 12) indicating high possibility of crash/collision if trip is started. For example, the safest route is same for the case trip start time 11AM and 4PM, however, this route is a lot safer at 11AM as compared to 4PM. This shows that our framework captures the impact of dynamic variables on safe route computation. Figure shows that different routes can be deemed to be safest at different times, moreover, safest route may not be desirable at all because it performs poorly in terms of safety score. Dynamic changes in variables impact safety route score and its corresponding safest routes, therefore, proposed framework enables users to make informed decisions on travel plans.

\section{Conclusion \& Future Work}
\label{conclusion}
We propose a safe routing framework called SafeRNet to addresses the safe transportation routing problem in the presence of Internet of Vehicles, cloud computing and Mobile Crowd Sensing technologies. The proposed framework addresses the need for computing the safest route and then delivers them back to interested users in a real time and on demand basis. User created and real time hardware device generated dynamic data are used to minimize the human errors. Bayesian network modeling approach and an optimization framework are used in cloud to analyze IoV and MCS generated spatio-temporal road traffic data. Furthermore, through experimentation on real data set we demonstrate that SafeRNet is effective in improving the transportation safety. 

There are limitations of our framework that we intend to work on in future. In our proposed framework, there needs to be a more structured approach to convert a map data into road network graph data for route computation purposes. We also believe a tradeoff between safest route and travel time/distance could be important to many users. We believe that this work would bring more attention to this important problem which holds the key to reducing the fatality/accident rates. 

\section*{Acknowledgment}
\label{Acknowledgment}
We thank the anonymous reviewers for their constructive feedbacks which helped us to improve this paper. We also like to thank National Institute of Statistics, Flanders (Belgium) for making the accident dataset available to the public.

\bibliographystyle{IEEEtran}
\bibliography{reference,bibfile_Kumar}

\end{document}